%% file: main.tex
\documentclass[letterpaper, 10 pt, conference]{ieeeconf}
\IEEEoverridecommandlockouts
\input{preamble.tex}

\title{\LARGE \bf Automated Pruning of Polyculture Plants}
\author{Mark Presten$^1$, Rishi Parikh$^1$, Shrey Aeron$^1$, Sandeep Mukherjee$^1$, Simeon Adebola$^1$ \\ Satvik Sharma$^1$, Mark Theis$^1$, Walter Teitelbaum$^2$, and Ken Goldberg$^1$
\thanks{$^1$The AUTOLab at UC Berkeley (automation.berkeley.edu)
{\tt\small \{mpresten, goldberg\} @berkeley.edu}
$^2$ Department of Robotics Engineering, UC Santa Cruz {\tt\small{wteitelb@ucsc.edu}}
}}

\begin{document}

\maketitle
\thispagestyle{empty}
\pagestyle{empty}

\begin{abstract}
\input{abstract}
\end{abstract}

\input{section-1-introduction}

\input{section-2-related-work}
\input{section-3-problem-statement}
\input{section-4-plant-phenotyping}
\input{section-5-bounding-disk}
\input{section-6-pruning-alg}
\input{section-7-pruning-hardware}

\input{section-8-experiments}
\input{section-9-conclusion}
\input{section-10-acknowledgements}

\nocite{goldberg2002beyond, harper1977population, toshioK}
\bibliographystyle{IEEEtran}
\bibliography{IEEEabrv,references}

\end{document}

%% file: preamble.tex
\usepackage[utf8]{inputenc}
\usepackage{graphicx}
\usepackage{multicol}
\usepackage{multirow}
\usepackage{array}
\usepackage{footmisc}
\usepackage{amssymb}
\usepackage{amsmath}
\usepackage{amsfonts}
\usepackage{algorithm}
\usepackage{algpseudocode}
\usepackage{tikz}
\usepackage{float}
\usetikzlibrary{math,shapes}
\usepackage{tabularx}
\usepackage{pbox}
\usepackage{ragged2e}
\usepackage{multirow}
\usepackage{balance}
\usepackage{subfig}
\usepackage{relsize}
\usepackage{hyperref}

\usepackage{listings}
\mathchardef\mhyphen="2D

\newcommand{\squeeze}{\vspace{-0.65cm}}

\usepackage{etoolbox}
\makeatletter
\patchcmd{\@makecaption}
  {\scshape}
  {}
  {}
  {}
\makeatletter
\patchcmd{\@makecaption}
  {\\}
  { -\ }
  {}
  {}
\makeatother
\usepackage{siunitx}

%% file: abstract.tex
Polyculture farming has environmental advantages but requires substantially more pruning than monoculture farming. We present novel hardware and algorithms for automated pruning. Using an overhead camera to collect data from a physical scale garden testbed, the autonomous system utilizes a learned Plant Phenotyping convolutional neural network and a Bounding Disk Tracking algorithm to evaluate the individual plant distribution and estimate the state of the garden each day. From this garden state, AlphaGardenSim \cite{avigal2020simulating} selects plants to autonomously prune. A trained neural network detects and targets specific prune points on the plant. Two custom-designed pruning tools, compatible with a FarmBot \cite{Farmbot} gantry system, are experimentally evaluated and execute autonomous cuts through controlled algorithms. We present results for four 60-day garden cycles. Results suggest the system can autonomously achieve 0.94 normalized plant diversity with pruning shears while maintaining an average canopy coverage of 0.84 by the end of the cycles. 
For code, videos, and datasets, see \href{https://sites.google.com/berkeley.edu/pruningpolyculture/home}{this url}.


%% file: section-1-introduction.tex
\section{Introduction}

Industrial agriculture is based on monoculture, where a single crop type is cultivated, requiring substantial use of fertilizer, pesticides, and water~\cite{rosenstock2014agriculture, AlbertoM}. Polyculture farming, on the other hand, is a sustainable practice in which multiple crop types are intermixed. The inherent benefits of polyculture farming include reduced weeds and soil erosion, better resistance to pests and viruses, and increased water and soil nutrient efficiency~\cite{risch1983intercropping, crews2018future, gliessman1982polyculture, liebman2018polyculture}. 
However, polyculture farming is more laborious than monoculture, requiring maintenance to ensure that larger, more dominant plant types do not overwhelm smaller, slow-growing plants. Furthermore, the inherent layout of polyculture farming makes non-invasive autonomous cultivation difficult due to the close proximity of each plant. 

\input{figures/disk_tracking}


We present an automated pruning system. We use an improved Plant Phenotyping and a Bounding Disk Tracking algorithm to predict individual plant centers and radii over time \cite{avigallearning}. A learned pruning policy identifies pruning actions to optimize plant diversity and coverage using the corresponding center and radii data. We introduce two novel, custom-designed pruning tools and algorithms that can autonomously prune plants alongside a learned Prune Point Identification network that identifies and selects specific leaves to prune. Experiments suggest that the autonomous system is capable of pruning plants to facilitate plant diversity while maintaining high canopy coverage. 
To the best of our knowledge, this is the first system in a polyculture farming setting capable of autonomously deciding and pruning plants. 

This paper makes five contributions:
\begin{enumerate}
    \item A Plant Phenotyping neural network that uses prior knowledge to predict plant types,
    \item K-Means and BFS-based Bounding Disk Tracking algorithms for garden state estimation, 
    \item Two automated pruning tool designs, 
    \item Evaluation of automated pruning hardware with a learned polyculture Prune Point Identification network,
    \item Experimental data from four 60-day Garden Cycles with analysis. 
\end{enumerate}

%% file: figures/disk_tracking.tex
\begin{figure}
    \vspace{0.2cm}
    \centering
    \frame{\includegraphics[clip, width=84.2mm]{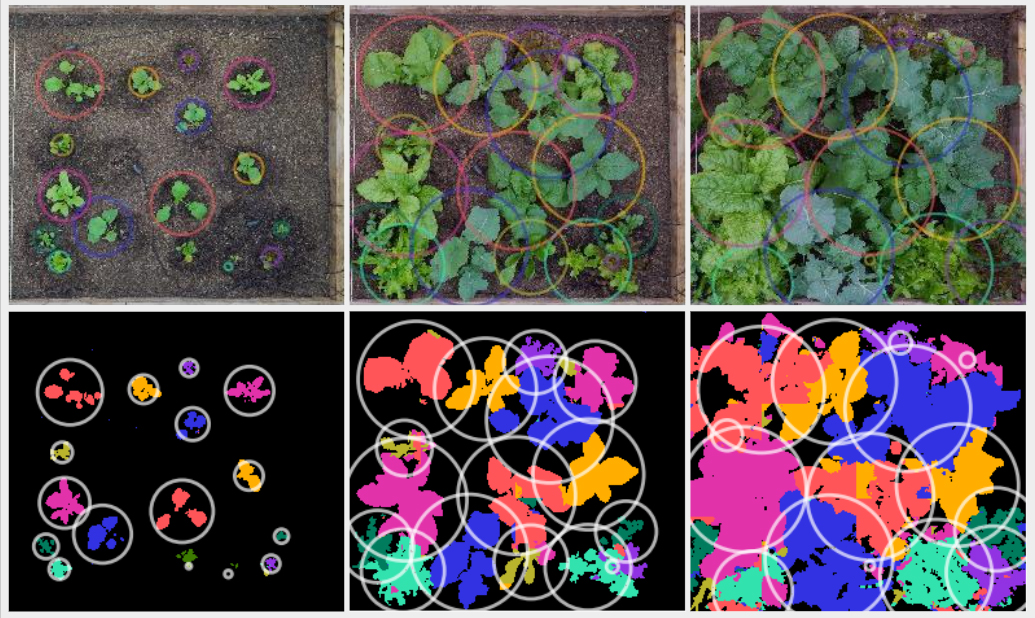}}
    \caption{\textbf{Phenotyping and Bounding Disk Tracking.} 3 images from days 20, 30, and 40 of garden cycle 4. \textbf{Top row:} overhead images overlayed with the estimated bounding disks from the Bounding Disk Tracking algorithm. \textbf{Bottom row:} the masks created by the Plant Phenotyping network as well as the estimated bounding disks (same as above). }
    \vspace{-0.7cm}
\label{fig:disk}
\end{figure}

%% file: section-2-related-work.tex
\section{Related Work}

Cultivating plants has been an essential human activity for over 10,000 years. Humans have continuously improved farming techniques, and in recent years, have introduced methods for agricultural automation. In 1995, the Telegarden, an art installation by Goldberg et al.~\cite{GoldbergK, goldberg2001robot}, allowed internet visitors to interact with a remote garden by planting and watering plants. Wiggert et al.~\cite{wiggert2019rapid} created a testbed for monitoring plant growth and water stress in real time. Our present work differs from these as we focus on autonomously pruning a diverse garden bed. 
Correll et al.~\cite{correll2009building} designed a distributed autonomous gardening system with mobile manipulators that detect plants, irrigate, and grasp fruits. While related, our work focuses on tools that would enable a fully automated polyculture pruning system.

\input{figures/autonomous_pipeline}

Existing plant simulators model growth of single species~\cite{stomph2020designing}. Examples such as DSSAT~\cite{jones2003dssat} and AquaCrop~\cite{steduto2009aquacrop} simulate a growth period in large-scale monoculture farms. In prior work, we presented AlphaGardenSim~\cite{avigal2020simulating, avigallearning, avigalsimulatingtase} - a polyculture garden simulator with first order models of plant growth, inter-plant dynamics, and competition for water and light. The model parameters used in AlphaGardenSim were tuned using real-world measurements from a physical garden testbed. AlphaGardenSim allowed us to simulate plant growth at an expedited rate and create automated policies and a seed placement algorithm optimized for plant diversity and coverage.   




Phenotyping is an important task for monitoring plants, similar to object tracking and identification. 
Ayalew et al.~\cite{leafcounting} present a method to use an unsupervised domain adaptation network to adapt the meticulously pre-labeled Computer Vision Problems in Plant Phenotyping (CVPPP) dataset~\cite{cvpppdataset, cvpppwebsite} to other plant and image domains. The data consists of single plants, their leaves, and a point map of leaf centers. This reduces the human effort required to track, count, and identify leaf centers. Our work builds on this work by transferring the results to a polyculture setting, as discussed in Section VI.


Pruning is a necessary capability to tend a polyculture garden. Prior work in autonomous pruning includes rose and bush trimming with a robot arm~\cite{cuevas, trimbot}. Habibie et al.~\cite{fruitmap17} trained a Simultaneous Localization and Mapping (SLAM) algorithm to enable automated fruit harvesting in a red apple tree field. Cuevas-Velasquez et al.~\cite{cuevas} demonstrated success using visual servoing to account for changes in stem poses to determine cutting points. 
In a controlled greenhouse, Van Henten et al.~\cite{Henten} used a robot with a thermal cutting tool to harvest cucumbers. We extend prior work by developing an autonomous pruning pipeline for trimming leaves in a controlled environment. To the best of our knowledge, this is the first case of autonomously pruning a polyculture garden. 

FarmBot is an open source gantry robot commercially available since 2016 that is used in our autonomous system. Prior work with this system has examined kinematic modeling to enhance FarmBot trajectory planning~\cite{farmbot1}. A team from Telkom University used FarmBot to create a web application to help human users with seed planting, watering, and plant monitoring routines~\cite{farmbot2}. More recently, researchers have proposed a FarmBot simulator ``to support the development of a control software able to implement different [precision agriculture] strategies"~\cite{farmbot3}. We use the FarmBot together with custom pruning and irrigation tools to tend a polyculture garden from planting, through germination, growth, reproduction and decay. 

%% file: figures/autonomous_pipeline.tex
\begin{figure}
    \vspace{0.2cm}
    \centering
    \frame{\includegraphics[width=80mm, height=55mm]{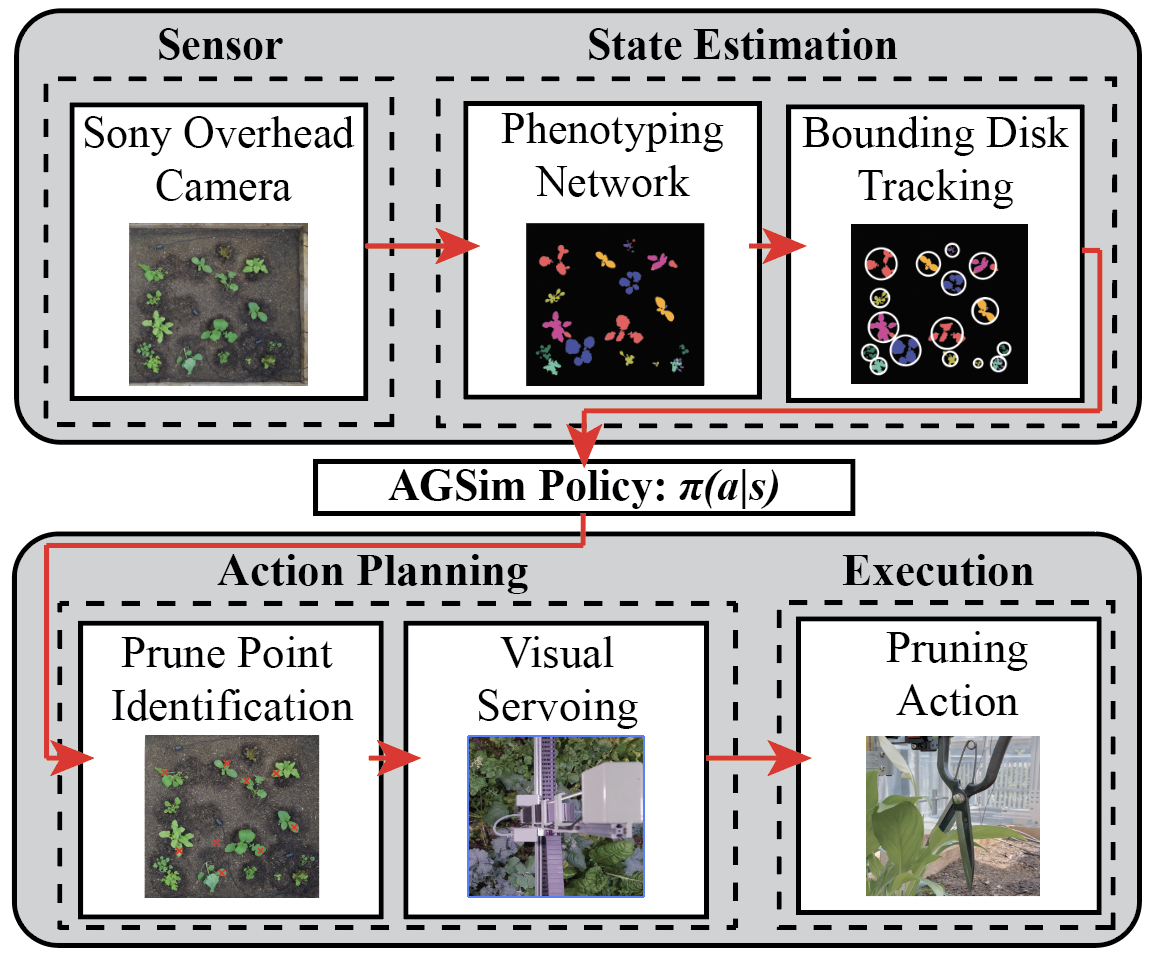}}
    \caption{\textbf{Automated Pruning Pipeline:} The overhead Sony camera takes photos on an hourly basis. The images are processed by a Plant Phenotyping Network followed by Bounding Disk Tracking algorithm to identify the garden's state. AlphaGardenSim determines which plants to prune in real time. Given the simulator's decisions, a Prune Point Identification network identifies specific leaves to prune. This is followed by visual servoing to arrive at the leaf location in the physical garden and then execution of the prune using a custom pruning tool.}
    \vspace{-0.5cm}
\label{fig:pipeline}
\end{figure}

%% file: section-3-problem-statement.tex
\section{The Polyculture Growing Problem}



A Garden Cycle consists of planting an arrangement of selected plant types, then irrigating and pruning until growth is completed. Garden quality is a function of coverage, plant diversity, and water usage. In this paper, we focus on maximizing coverage and diversity through pruning actions.

Each Garden has a total of $n$ plants, placed within a planter bed of size $(w, h)$ in $cm$. For each plant $k \in [0, n)$, the plant has its center coordinates $(cx_k, cy_k)$ and current radius $r_k$, both in $cm$. Each plant $k$ also has a corresponding plant type $i$ (equivalent to $p_k$), which dictates the estimated germination time $g_i$, maturation time $m_i$, and maximum radius $R_i$. The lifecycle of each plant $k$ is defined by four stages: germination, growth $l_i := m_i - g_i$, waiting, and wilting, as defined in \cite{avigal2020simulating}. A garden state on day $t$ includes all information described above for every plant $k \in [0, n)$ at day $t$. Thus, a garden state is defined as follows:
$$s(t) = \{ p_k : ((cx_k, cy_k), r_k), ... \}, k \in [0, n)$$


We define coverage, $c(t)$, as the sum of all plant type canopy coverage, $\Sigma_i c_i(t)$, over total area $w \cdot h$ at day $t$. For each plant type $i$ with maximum radius $R_i$, we define garden diversity as follows:
\begin{align*}
    &\textbf{v}(t) = [c_{i}(t) \cdot (R/R_i)^2, ...], \forall i\\ &d(t) = H(\textbf{v}(t))
\end{align*}
where $H(\cdot)$ is an entropy function, $\textbf{v}(t)$ is a vector of normalized plant type coverage, and $R$ is the average maximum radius over all plant types. Multiplying $c_{i}(t)$ by $(R/R_i)^2$ normalizes each plant type's canopy coverage. We normalize because it is unrealistic to assume that a smaller, less dominant species will have the same coverage as a much larger, faster growing species. We aim to maximize both $c(t)$ and $d(t)$ through pruning actions. 

The physical plants are grown in a $3.0m\times1.5m$ raised planter bed located in the UC Berkeley greenhouse. We split the planter bed into two halves and grow identical seed placements ($1.5m\times1.5m$) in each with different pruning regiments (see Section VIII). The cycles last 60 days.




%% file: section-4-plant-phenotyping.tex
\section{Plant Phenotyping}

To estimate the garden state, we use a learned semantic segmentation neural network to label plant types from an overhead image. Plant phenotyping directly influences the success of Bounding Disk Tracking (Section~\ref{boundtrack}), and provides information on plant growth, diversity, and coverage. 

We mounted a Sony SNC-VB770 digital camera~\cite{sonycamera} with a 20$mm$ Sony lens~\cite{sonylens} 2$m$ above the garden bed to monitor the garden. The VB770 satisfies our major requirements that include (1) resolution, (2) image distortion, (3) FOV, (4) power delivery, and (5) remote data accessibility. It has a DSLM 35$mm$ sensor with a maximum 4240$\times$2832 resolution, and the camera publishes photos every hour. We trained a model using UNet architecture ~\cite{unet_architecture} and ResNet34~\cite{resnet_architecture} backbone to output a $1630\times3478\times (i_{total} + 1)$ array $L$ of plant likelihood per pixel per label type, where $i_{total}$ is the total number of plant types. The network is trained on six hand-labeled overhead images from previous garden cycles. Each image is split into $512\times512$ RGB patches and augmented via shifting and rotating. We extract leaf masks from various stages in the garden and overlay these leaves on top of the existing patches to augment the data set ~\cite{avigallearning}. 

Hand labeling accurate ground truth masks is a tedious process. We developed a data aggregation based approach, allowing a human to make corrections to a predicted mask when the algorithm fails. This approach identifies plant sub regions using the contours of the prediction mask, and queries a human to generate the correct label. This method allowed us to quickly generate training data from multiple garden cycles to improve overall performance. 

Accurate segmentation for plants after day 30 becomes increasingly important to be able to determine canopy coverage and pruning actions. However, a plant may look very different at germination compared to its mature state due to the distribution shift of a plant over its lifespan (as well as due to occlusions), which causes a drop in performance starting on day 40. 

To address this, we introduce a prior probability distribution based on seed placement and plant maximum radius given from our tuned simulator \cite{avigal2020simulating}. We define a variable $R_t^k$, and $c_t^k$ as the maximum radius and center of plant $k$ at timestep $t$, and a $1630\times3478\times (i_{total} + 1)$ occupancy grid, $O$ defined as: $$O(x, y, i) = \alpha * (2 - r / R_k)$$ if $r \le R_k$ and $c_k$ is of plant type $i$, where $\alpha = 5$, and $r$ is the distance from $c_k$ to $(x, y)$, else 1.


We use this location based occupancy grid as a prior probability, and compute a new likelihood grid  $L^\prime$ as an element-wise multiplication of the original segmentation output, $L$, and occupancy grid, $O$, $L^\prime(x, y, i) = L(x, y, i) * O(x, y, i)$, and output $\max_i L^\prime(x, y, i)$ as the predicted label for $(x, y)$. 

We define mean IoU as $\sum_{i=0}^{i_{total}}{IoU(\text{label}_i)} / (i_{total} + 1)$. The baseline model ~\cite{avigallearning} had a mean IoU of 0.71 when compared to the ground truth at day 30. The new network, with data aggregation techniques and location based segmentation added, had a mean IoU 0.83 across the 9 labels on day 30. We saw the highest IoU of 0.97 in borage, which is one of the larger plants. Radicchio, which previously had the lowest IoU, had the largest increase from 0.23 to 0.59.


Adding location priors offers more robustness to the distribution shift in plants towards the end of the garden cycle and marginal improvements in the early stages of the garden. During day 50 and 60, mean IoU improved from 0.38 and 0.33 to 0.42 and 0.36 respectively with location based segmentation. The largest jump in IoU was for  green lettuce, from 0.31 to 0.40 on day 60, while plants like kale saw little change with an IoU of 0.54 on both networks.

%% file: section-5-bounding-disk.tex
\section{Bounding Disk Tracking}\label{boundtrack}



Tracking plants over the lifespan of a garden is challenging because plants' shapes frequently change day-to-day due to occlusion, but bounding disks should remain consistent. 

We define a plant's bounding disk (see Fig.~\ref{fig:disk}) as the circle with the smallest radius such that all pixels corresponding to that plant are enclosed. This definition helps account for plants moving over time due to phototrophy~\cite{whippo2006phototropism} and irrigation~\cite{dietrich2018hydrotropism}. We present two methods for finding circular representations of the garden's state and two metrics for comparison, and evaluate each method against a hand-labeled benchmark for selected days using a circle IoU loss~\cite{DBLP:journals/corr/abs-2006-02474}.

To estimate the garden state, defined by plant centers and radii ($(cx_k, cy_k), r_k$) indexed by plant type $p_k = i$, we convert the plant segmentation mask into estimates of each plant's center and radius. It is necessary to phenotype the overhead image before converting from real-life (real) to simulation (sim) to ensure pixels with the highest likelihood for that plant type affect its bounding disk representation. 

We use a breadth-first-search (BFS) based algorithm and K-Means clustering to track each plant’s center and radius. Both algorithms help address the issues with tracking plants over the duration of the garden lifecycle. BFS helps with irregular plant shapes and slight occlusions by continually searching outwards using a radial search heuristic, and K-Means helps address occlusion because it clusters non-contingent groups of pixels into a single bounding disk. 

The BFS algorithm is initialized with seed locations and all plant radii at 0$cm$. At each timestep, we use AlphaGardenSim~\cite{avigal2020simulating} and the prior plant radius to calculate a maximum possible radius by simulating a day of plant growth. Given the prior radius, maximum radius, and minimum radius, the algorithm traverses outwards from the minimum radius. The algorithm stops when less than 10\% of the newly traversed pixels are of the correct type or the maximum radius has been achieved. This process repeats each day for each plant. Even when a plant becomes fully occluded, the algorithm handles radial decrease using AlphaGardenSim’s tuned wilting parameters.

The second method is a K-Means clustering based algorithm. K-Means clustering has two main assumptions -- that the clusters (1) have roughly the same number of points and (2) are circularly distributed. The first assumption is true near the beginning of the garden, because plants of the same type grow similarly. However, later in the cycle, competitive relationships in the garden and occlusion start to create asymmetries, complicating this assumption. The second assumption follows from the circular model we use to track plants. 


In order to benchmark the performance of these methods, we introduce two metrics: average circle utility (ACU) and percentage of pixels included (PPI). Each of these metrics is computed per plant type per timestep. Let $P_i$ be the number of pixels in the segmentation mask of the inputted plant type that fall within at least one bounding disk, $P_t$ be the number of pixels of the given plant type present in the segmentation mask, and $P_c$ be the area of the union of the fitted bounding disks. We define the average circle utility as $\text{ACU} = \frac{P_i}{P_c}$ and percentage of pixels included as $\text{PPI} = \frac{P_i}{P_t}.$ 

We want to maximize both of these metrics, ACU and PPI, to compute the optimal bounding disks. On the extremes, these algorithms are adversarially related; smaller bounding disks tend to have higher ACUs because they will likely be centered around denser, less occluded portions of the plants. However, larger bounding disks will tend to have higher PPIs because a larger bounding disk will naturally have a larger portion of a plant $k$'s pixels. 

\input{figures/garden_metrics}

To judge the efficacy of these methods we compare them to hand-labeled bounding disks at various time steps. As Fig.~\ref{fig:metrics} (left) shows, initial K-Means clustering performs well as its assumptions are easily met and the segmentation is highly effective. It also performs well on larger, less occluded plants. However, later in the cycle, this method's efficacy decreases as it overfits to segmentation errors and irregular plant shapes. As Fig.~\ref{fig:metrics} (right) shows, BFS lags early on, but then becomes increasingly effective as plants are occluded mid-garden cycle.

%% file: figures/garden_metrics.tex
\begin{figure}
    \vspace{0.2cm}
    \centering
    \frame{\includegraphics[width=4cm]{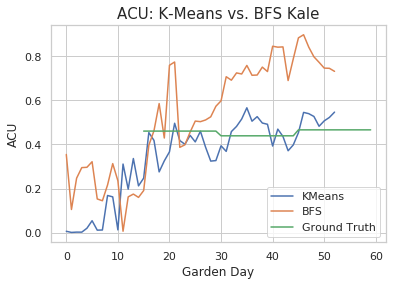}\includegraphics[width=4cm]{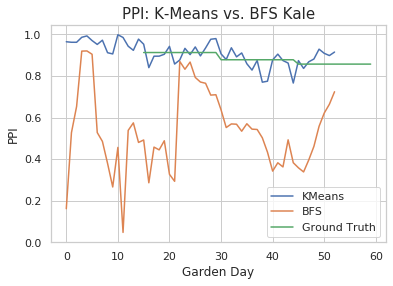}}
    \frame{\includegraphics[width=4cm]{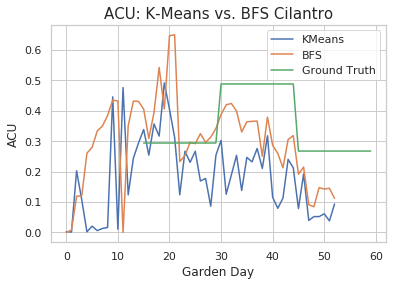}\includegraphics[width=4cm]{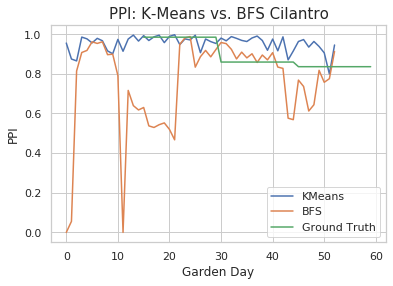}}
    \caption{\textbf{Garden Metrics of Garden Cycle 2R for Kale and Cilantro}. We evaluate average circle utility (ACU) and percentage of pixels included (PPI) of the Breadth-First-Search (BFS) versus the K-Means bounding disk algorithms for Kale, a larger plant type, and Cilantro, a smaller plant type.
    \textbf{Kale:} BFS tends to have higher ACU, but lower PPI. For the days which ground truth circles exist (manually annotated), they are closer to the K-Means algorithm in both metrics. \textbf{Cilantro:} Similarly, BFS has a higher ACU and K-Means has a higher PPI. However, Cilantro generally benefits from the more conservative BFS. 
    We adopt a mixed approach: the K-Means approach for larger plants and less occluded timesteps, and the BFS approach for denser, smaller plants.
    }
    \vspace{-0.7cm}
\label{fig:metrics}
\end{figure}

%% file: section-6-pruning-alg.tex
\section{Pruning Planner}

Once a garden state for day $t$ is estimated with the Bounding Disk Tracking algorithm, the analytic policy within AlphaGardenSim decides which plants to prune. For autonomous pruning, the system must identify and select specific target leaves to prune, be able to navigate and position the FarmBot above the chosen leaf using visual servoing, and execute the pruning action with custom hardware. 


\subsection{Prune Point Identification}

The system must identify the best leaf to cut after a plant is chosen to be pruned by AlphaGardenSim. Our baseline approach found the average point between an extrema of the plant, a point near the tip of a leaf as dictated by the bounding disk, and the plant center to find a theoretical leaf center. However, this was constrained by the reality of plants’ physical makeup which often includes bending, occlusion, or oddly shaped leaves. The algorithm would frequently return points which were not on a plant or too close to an edge. We therefore explore a learned approach. 

We trained a Prune Point Identification neural network based on the unsupervised domain adaptation network for plant organ counting by Ayalew et al. \cite{leafcounting}. In the training process, our images are transformed to match the input network characteristics, allowing for a more seamless domain adaptation. The architecture consists of a Domain Adversarial Neural Network with a Gradient Reversal Layer to backpropagate between the source and target domains and classification is performed using a U-Net \cite{leafcounting}.

To evaluate this network's success in a polyculture setting, rather than its original monoculture domain, we trained it on all plant types, different sets of plant types that appeared to have distinct leaves, and on individual plant types from our domain.
We found that training on all plant types led to the worst overall performance. Borage, a plant that has high success in being identified by our phenotyping network along with distinct, well-shaped round leaves, led to a network that was best able to predict leaf centers for all plant types. The final model was trained for 150 epochs with a 80/20 train/validation split for the source (CVPPP) and target datasets, 201 overhead images and masks of the Borage plant type, and evaluated visually on a random sampling of overhead images of all plant types. 

The model generates a heatmap with all possible plant leaf centers. A clustering and thresholding technique is used to identify leaf centers with the highest model confidence. These points are then removed and the heatmap is re-normalized to identify less certain points. The algorithm is able to recover lower confidence leaf centers, compared to the initial normalized threshold of 0.3, while accounting for over-classification. The algorithm ensures that prune points do not land on other plants or the soil through the use of the phenotyping mask.
Together, the model and recursive algorithm identify 32\% more leaf centers than the baseline methodology (see Fig.~\ref{fig:oh_keypt}). The center of mass for the identified points is an average of 38\% closer to the center compared to the baseline. Pruning closer to the center of the plant is beneficial because it allows for pruning actions to cut off a greater portion of the leaf. Furthermore, as seen in Fig.~\ref{fig:oh_keypt}, the learned method has far fewer points that lie on different plants. 

For prune point selection, the network first identifies all possible prune points. The algorithm then eliminates all points within $3cm$ of the edge of the bounding disk, and calculates the rate of change of the radii of all neighboring plants over the last five days. The prune point that is closest to the neighboring plant that has the largest rate of decay of radii is selected in order to foster growth of the struggling neighboring plant. 

\input{figures/overhead_keypoints}



\subsection{Visual Servoing for Pruning Tool}

The autonomous system must then physically arrive at the chosen prune point by translating from overhead image pixel coordinates to FarmBot $(x, y)$ coordinates. Due to the variable height of plants, it is not possible to create a 1-to-1 mapping of pixel coordinates to FarmBot coordinates. 

The visual servoing algorithm works using an on-board snake inspection camera located adjacent to the tool end effector on the FarmBot Z-axis \cite{raspicamera}. It allows for close-up images of plants and soil. Given plant $k$ was chosen to be pruned, the FarmBot moves to its original seed location and takes a photo using the on-board camera. This image is then localized within the overhead `global' image by calculating a normalized correlation coefficient between the images. Instead of exhaustively searching the entire garden bed to localize the image, the servoing algorithm constrains the search to a max area around the prune plant's center within the global image, dictated by the FOV of the on-board camera. The algorithm also iteratively tests different scales of the on-board image, which accounts for the variable height of the canopy, and finds the scale and position that has the highest coefficient. 

After finding the best match in the overhead global image, the FarmBot is instructed to move along the vector from the current location to the prune point. Then an iterative process begins, in which a `local' image is taken at the new point and is localized within the global image. Once localized, the FarmBot moves in the vector direction a max distance of $4cm$ to prevent erroneous movement if a local image is miss-classified within the global image. The iterative cycle continues until the FarmBot reaches within $1cm$ of the prune point or reaches an iteration limit of six.

Although the visual servoing algorithm is quite robust, when plants grew too high and close to the on-board camera (approximately $0.4m$ above the soil surface), it was not able to take a clear image of the garden, which led to failed localization and servoing to the prune point location. To remedy this, we moved the on-board camera to approximately $0.7m$ above the soil surface, away from plants' reach, allowing it to capture unobstructed images of the garden and better localize them within the global image.



%% file: figures/overhead_keypoints.tex
\begin{figure}
    \vspace{0.2cm}
    \centering
    \frame{\includegraphics[width=84mm, clip]{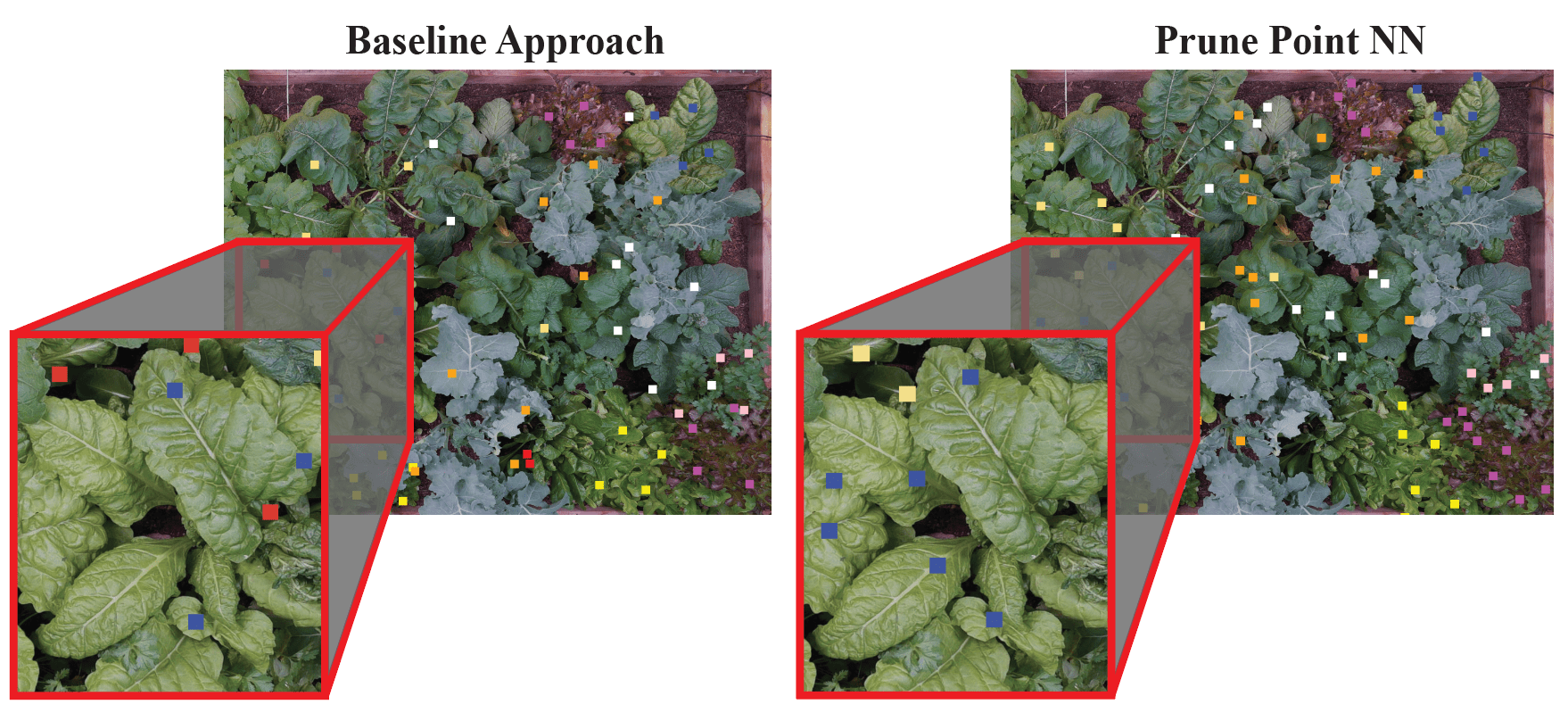}}
    \caption{\textbf{Prune Point Identification.} Example of all plant leaf centers that were identified by the baseline algorithm (left) and the learned model (right) applied to an overhead image. Each prune point color corresponds to a different plant type. The learned model identifies more usable points with fewer misclassifications. When looking at the Swiss Chard plant (zoomed in), we see that the learned model finds 3 more prune points than the baseline approach and also does not missclassify the red prune point, which is meant for a neighboring plant type.}
    \label{fig:oh_keypt}
    \vspace{-0.7cm}
\end{figure}

%% file: section-7-pruning-hardware.tex
\section{Pruning Hardware}

\input{figures/pruner}

The goal of pruning is to reduce the coverage of plant $k$ centered at a point $(x_k,y_k)$ with radius $r_k$. To tend to and prune plants, the autonomous system uses a commercial FarmBot~\cite{Farmbot} installed over the $3.0m\times1.5m$ planter bed frame. This CNC robot can travel to any location in the garden from the soil level to 0.4$m$ above. The FarmBot also features a magnetic universal tool mount (UTM) on its Z-axis that can automatically swap between tools stored on the west side of the bed. The tools we designed are operated through the FarmBot system with no human intervention. Once the FarmBot moves to a prune location, the pruning tool then aims to remove all or part of the leaf structure in that neighborhood to reduce coverage. We designed, implemented, and studied two options.

\paragraph{Rotary Pruner}
We built a custom pruning tool, dubbed the Rotary Pruner, that is lightweight, integrates with the FarmBot universal tool mount, and mounts automatically. Inspired by the traditional `weed whacker,` our first generation model utilizes thin, flexible blades rotating at high speeds to cut plants. We selected an SM Tech 775 Brushed 24V DC motor capable of 12000 rpm to achieve this. The motor's high power needs ($>$5V) mandated an external voltage source separate from the Farmbot's power rail. Thus, we designed a spring pin mechanism that allows the external power rail to automatically connect to the tool. We also designed a motor housing that inter-operates with the FarmBot UTM. The electrical control includes a relay circuit that governs motor power and uses GPIO to integrate with the FarmBot OS. The FarmBot does not rotate along the Z-axis, so we designed two such rotary pruning tools with different orientations: one that cuts along the X-axis and another that cuts along the Y-axis.




The Rotary Pruner that is chosen has a cutting direction that is closest to being orthogonal to the vector from the plant's center to the prune point, and is autonomously mounted using the tool rack and FarmBot UTM. To estimate the height of the plant and find the distance to the target leaf $d$, we mounted a Sharp infrared distance sensor \cite{depthsensor} adjacent to the FarmBot UTM pointing towards the soil surface. After arriving at the prune coordinates and measuring $d$, the Rotary Pruner is then toggled on, and the FarmBot is lowered to $d+5 cm$; the system overestimates the depth of the leaf in order to ensure a cut. The Rotary Pruner is then toggled off and returned to its home position. 

The Rotary Pruner faced fundamental limitations, primarily with its aggressive method of operation (the high speed blades would cause debris to fly), which could pose a danger to objects and people around the garden. 




\paragraph{Pruning Shears}

Although the Rotary Pruner proved useful for many of the initial pruning actions, it spotlighted a few shortcomings that we wished to fix with a redesigned pruning attachment. Firstly, since the Rotary Pruner uses two separate attachments, the autonomous system had to regularly switch these attachments, adding unwanted power consumption and increasing the likelihood of mechanical failure. Secondly, due to the Rotary Pruner's relatively aggressive method of operation, it would frequently damage the target leaf (as well as surrounding plants) when attempting a prune action. This caused a reduction in plant health and an increase in water consumption.

For a quieter, more precise and delicate pruning tool, we motorized a pair of Japanese topiary shears. A pair of Niwaki Topiary Shears~\cite{niwakishears} 
were fastened directly to the FarmBot's gantry rails. A YANSHON Digital 360$^\circ$ servo motor is able to close the shears by winding a high strength steel cable attached to one handle of the shears onto a spool; the shears reopen with a spring mechanism when the cable is unwound. This assembly is mounted to a 2-axis servo gimbal (using BETU Digital 270$^\circ$ servo motors). The gimbal is able to position the shears vertically, horizontally, or at any intermediate angle as well as rotate the shears a full 180$^\circ$ to account for any leaf direction, allowing the FarmBot to trim with greater precision as well as reach the tops of plants. The servos connect to the FarmBot PWM header and integrate seamlessly with the FarmBot OS.

Control of the shears is executed through the three servos: one for tilt, one for cut angle, and one for shear closure. The Pruning Shears are at default open and stored horizontally to avoid collisions with plants below. The shears require calculating the orthogonal vector to the vector spanning from the center of the plant to the prune point. The servo that controls cut angle is then activated to position the shears along the orthogonal vector. The tilt servo then swivels the shears to a vertical position. The shears are then lowered to $d+5 cm$ and activated. Once a cut is complete, the shears return to their default positioning.  


\input{figures/isolated_prunes}

%% file: figures/pruner.tex
\begin{figure}
    \vspace{0.2cm}
    \centering
    \frame{\includegraphics[height=35mm, width=42mm]{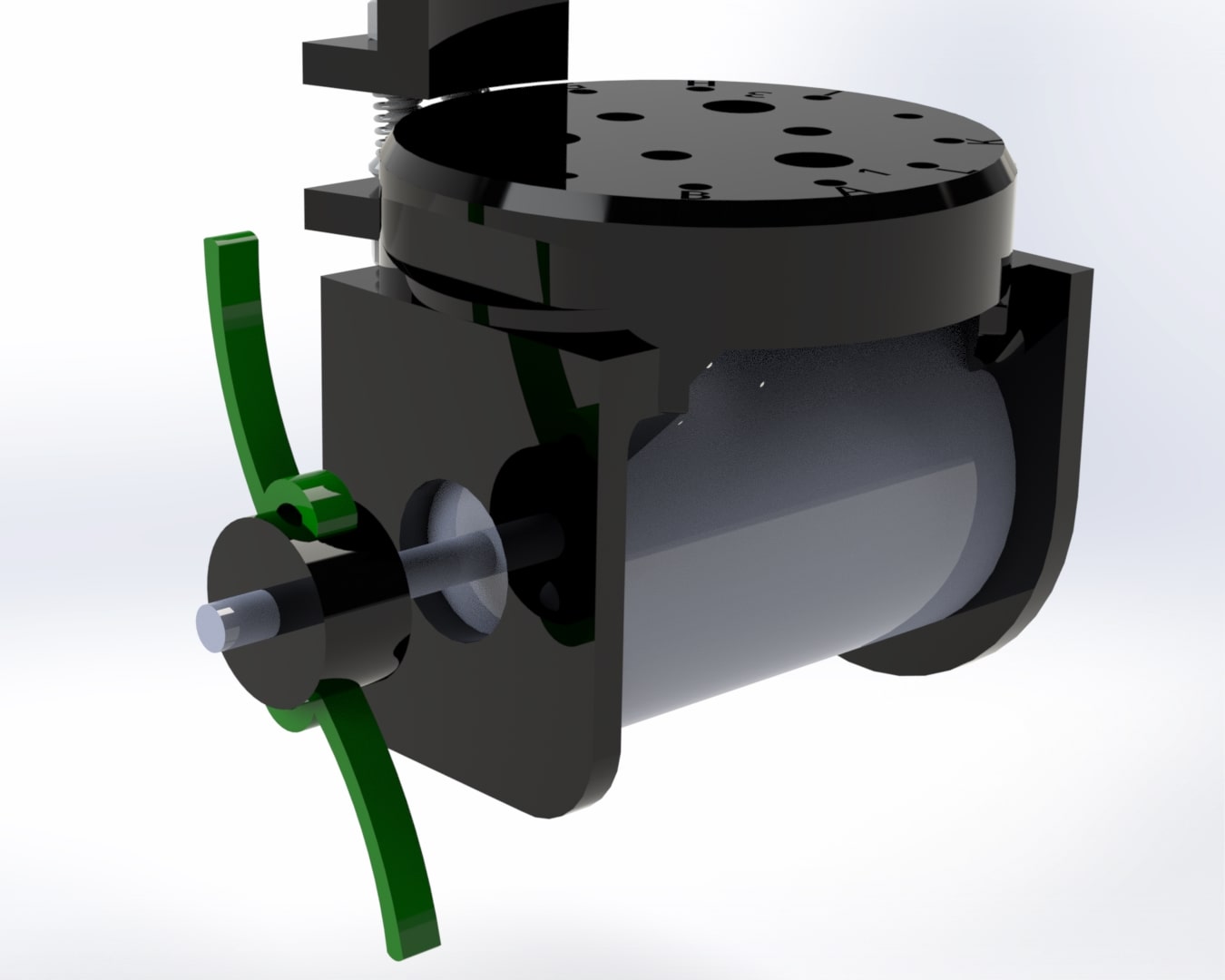}}
    \frame{\includegraphics[height=35mm, width=42mm]{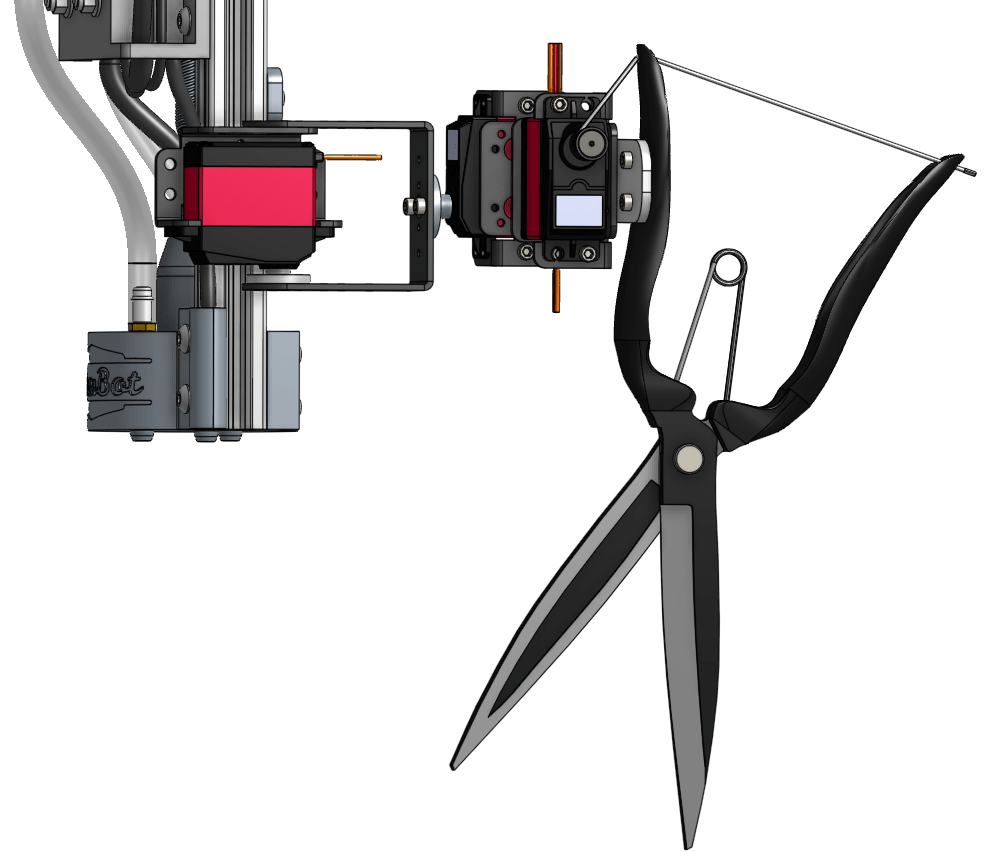}}
    \frame{\includegraphics[height=35mm, width=42mm]{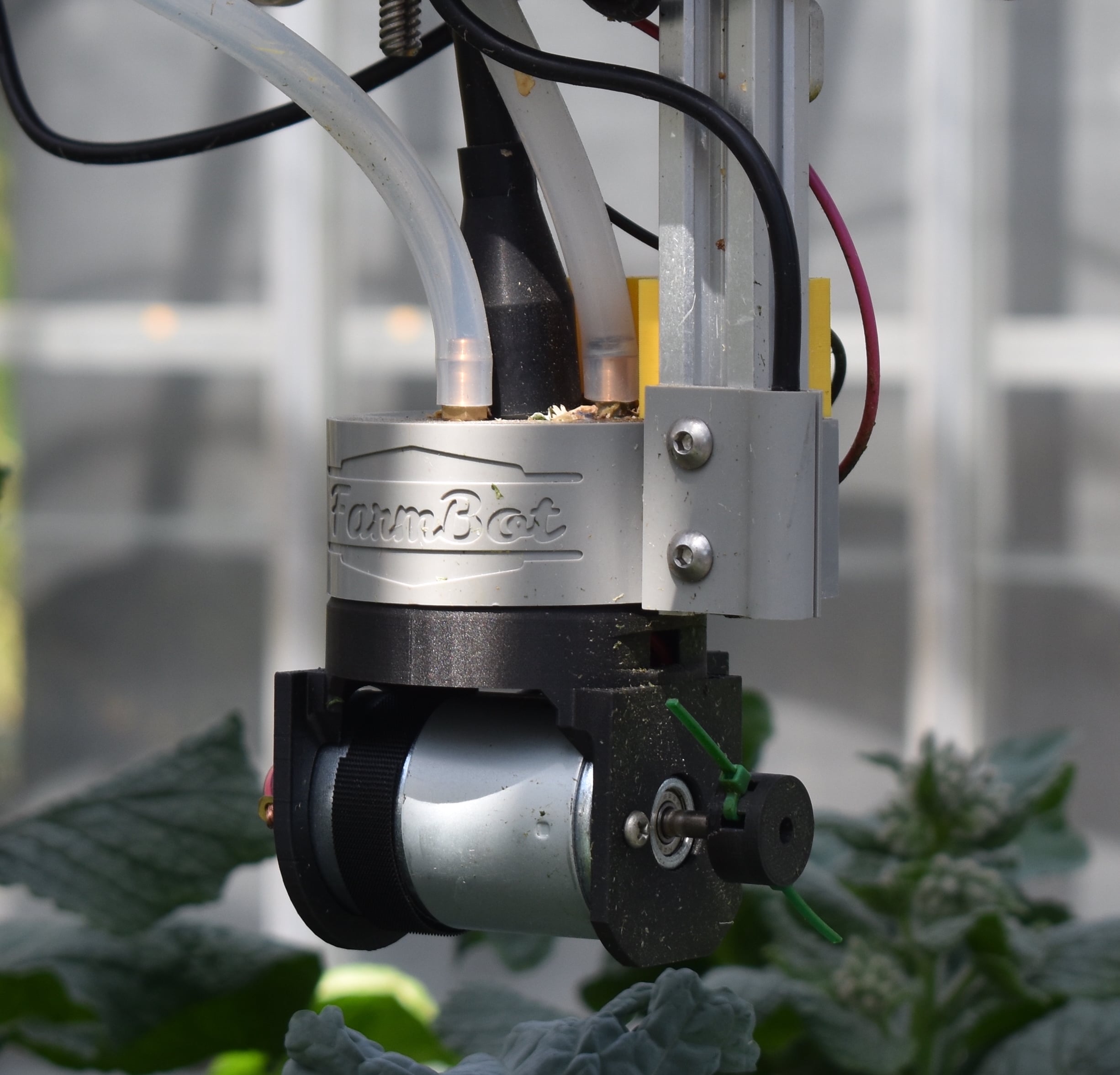}}
    \frame{\includegraphics[height=35mm, width=42mm]{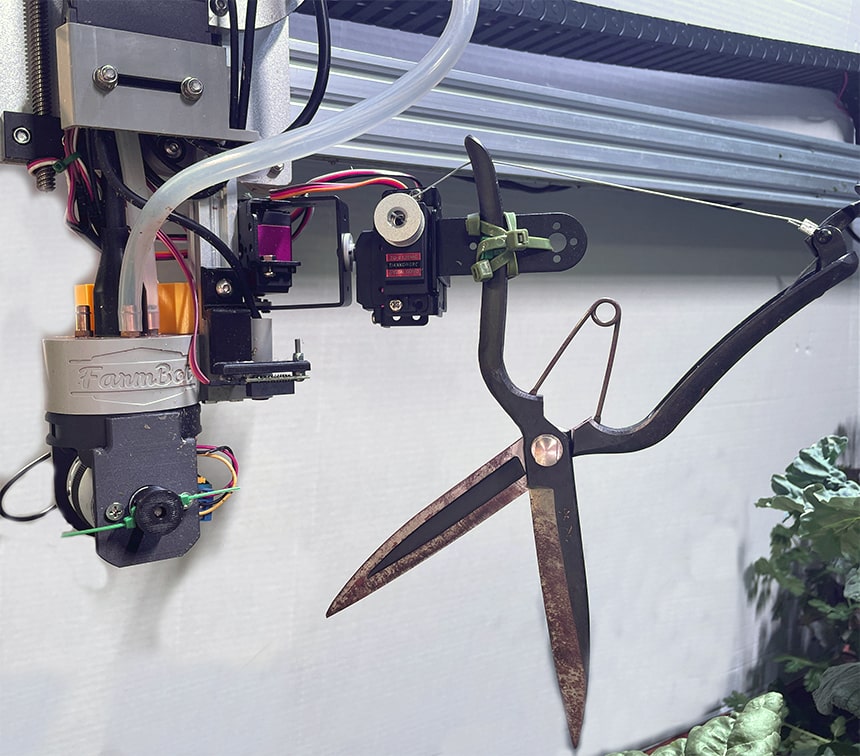}}
    \caption{\textbf{Pruning tools.} \textbf{Left:} CAD and physical model of Rotary Pruner with a high speed motor and trimming blades. \textbf{Right:} CAD and physical model of Pruning Shears with three servos to control closing, tilt, and orientation.}
    \vspace{-0.7cm}
\label{fig:sci_pruner}
\end{figure}

%% file: figures/isolated_prunes.tex
\begin{table}
\centering
\vspace{0.15cm}
\scalebox{0.85}{
\begin{tabular}{|c|c|c|c|c|c|c|c|}
\hline
Plant Type & Cut & \multicolumn{3}{|c|}{Rotary Pruner Results} & \multicolumn{3}{|c|}{Pruning Shears Results}\\
\hline
 & & Compl. & Precision & Err. & Compl. & Precision & Err. \\
\hline
\hline
Eggplant & 1 & 2 & 0 & B & 2 & 0 & B\\
\hline
 & 2 & 2 & 1 & A & 3 & 0 & A\\
\hline
 & 3 & 3 & 1 & A & 2 & 0 & B\\
\hline
 & 4 & 3 & 1 & A & 3 & 1 & A\\
\hline
 & 5a & 2 & 0 & B & 2& 0 & C,D\\
\hline
 & 5b & 2 & 0 & B & 3 & 0 & C,D\\
\hline 
 & 6a & 2 & 1 & B & 2 & 1 & D\\
\hline 
 & 6b & 2 & 1 & B & - & - & -\\
\hline
\hline
 BellPepper & 1 & 1 & 1 & D & 1 & 0 & B\\
\hline
 & 2 & 1 & 1 & B & 2 & 0 & C\\
 \hline
 & 3 & 1 & 1 & B & 3 & 0 & A\\
 \hline 
 & 4 & 3 & 0 & A & 1 & 0 & B\\
 \hline
 & 5a & 3 & 0 & A & 1 & 0 & C,D\\
 \hline
 & 5b-c & - & - & - & 1 & 0 & C,D\\
 \hline
 & 5d & - & - & - & 3& 1 & C,D\\
 \hline
 & 6 & 2 & 1 & B & - & - & -\\
 \hline
\end{tabular}}
\caption{\textbf{Isolated Pruning Experiments} for the Rotary Pruner and Pruning Shears. \textbf{Key:} \emph{Completeness}- 3: complete cut, 2: partial cut, 1: missed cut. \emph{Precision}- 1: no damage to other leaves, 0: damage to other leaves. \emph{Error Type}- A: No error, B: location, C: depth, D: Other.}
\squeeze
\vspace{-0.2cm}
\label{table:isolated_prunes_shears}
\end{table}

%% file: section-8-experiments.tex
\section{Experiments}

\subsection{Isolated Pruning Experiments}

To evaluate the two pruning tools, we ran isolated pruning experiments using two plant types in pots. We chose eggplant for its large leaves comparable to kale, borage, and turnip, and we chose bell pepper for its smaller abundant leaves similar to cilantro and lettuce. We placed a grown potted plant near the midline of the garden bed, took an overhead image, and then passed a manually annotated plant center and prune point into our visual servoing and pruning algorithms.

For each tool, we made 5-6 prunings on both plant types, observing completeness of the cut, precision of the cut (if any neighboring leaves were harmed in the process), and any error that may have occurred. We found the Rotary Pruner was more likely to complete a cut, but it also tended to over prune. Furthermore, due to the nature of the Rotary Pruner, the final cuts were not 'clean' and showcased tears and fragments. The Shears, on the other hand, generally caused little secondary damage and debris and made clean cuts, but were more prone to incompletely cut or miss a leaf. Table~\ref{table:isolated_prunes_shears} shows the results. The main reasons for failure to execute a prune were due to bad prune point selection or the pruning tool pushing a leaf out of the way. 



\subsection{4 Garden Cycles}
To evaluate the entire system holistically, we ran four autonomous cycles over two 60 day periods. We split the garden into two halves and planted identical seed placements ($1.5m\times1.5m$) on each. Each half was treated as an independent garden cycle. Irrigation took place at 9:00 AM daily and every plant was watered 200$mL$. After day 30, and every five days after, the autonomous system executed pruning actions. An overhead image taken at 7:00 PM was processed through the Plant Phenotyping and Bounding Disk tracking algorithm to determine the garden state. AlphaGardenSim would use this garden state to decide which plants to prune. The image was subsequently used for prune point identification and selection. Visual servoing and pruning algorithms were then executed on the chosen leaves.

\paragraph{Human Intervention}

Although the goal is a fully automated polyculture garden pruning system, some human intervention was required during the Garden Cycles. Seed planting was performed with human labor. A member of the project team (Mark) was present during all pruning actions, which were executed in batches. While all decisions were made autonomously, human intervention was used to correct robot position when the FarmBot gantry failed to servo to the correct target location, this occurred in 45\% of pruning operations. However, by adjusting the on-board camera height, we were able to reduce the rate of fail of visual servoing for future garden cycles. No other human intervention was performed in terms of weeding or irrigation.

\paragraph{Garden Cycles 1L and 1R}
In Cycles 1L and 1R, the identical seed placements ($1.5m\times1.5m$) included 20 plants from 10 different plant types (two of each type). In Cycle 1L (the left half of the garden bed) there were no pruning actions and the garden was allowed to grow freely. In Cycle 1R (the right half) pruning actions were executed with the Rotary Pruner. 

Over 6 pruning sessions for Cycle 1R, 42 plants were chosen to be pruned across 6 plant types. The system autonomously selected the turnip and kale plants on all pruning occasions, most likely due to the fact that they grew much faster than the other plants and have large radii. Due to the numerous prunings and the Rotary Pruner's nature of completing a cut and leaving a leaf vulnerable, we see both turnip plants approach their wilting stage by day 60. This could also be a sign of overpruning. 

Final canopy coverage and diversity are reported in Table~\ref{table:cycle_1_div} for each individual plant type. The metrics were found through creating a manually labeled ground truth mask on day 60 of the garden cycle. As seen by comparing the results with and without pruning, it is clear that pruning increases diversity by creating space for smaller plants to develop. The larger plants coverage decreased while the smaller plants coverage increased, leading to a more diverse garden overall (13.32\% increase). This increase in diversity did come at the cost of losing some overall coverage (15.15\% decrease). 

\input{figures/cycle_1_diversity_and_coverage}

\paragraph{Garden Cycles 2L and 2R}
For Garden Cycles 2L (left) and 2R (right), we planted two identical seed placements ($1.5m\times1.5m$). 
Cycles 2L and 2R included only 16 total plants from 8 plant types. Sorrel and arugula were omitted as sorrel was relatively much smaller than other plants in the garden and arugula had the tendency to grow too tall, impeding movement of the FarmBot gantry system. 

For Cycles 2L and 2R, all pruning actions were performed using the Pruning Shears, and, as before, the two halves were treated independently. During Cycle 2L, 35 plants were chosen for pruning across 6 plant types, while during Cycle 2R, 38 plants were chosen across 7 plant types. We see a decline in the total number of prunings compared to Cycle 1R because of the fewer number of plants in the garden. Kale and borage (two of the largest plants in the garden) were most commonly selected in both garden cycles. No plants exhibited signs of wilting or overpruning by day 60. 

\input{figures/cycle_2}

To evaluate Garden Cycles 2L and 2R relative to Cycles 1L and 1R, we manually created segmentation masks for days 20, 30, 40, 50, and 60. Fig.~\ref{fig:cycle_2} shows coverage and diversity graphs for all four garden cycles. We found the autonomous system to achieve an average of 0.94 normalized diversity with the Pruning Shears for Cycles 2L and 2R on day 60, and an average canopy coverage of 0.84. While the Rotary Pruner exhibited a higher diversity metric (0.97), the Pruning Shears outperformed the non-pruned garden, Cycle 1L, in terms of diversity (0.85) while sacrificing much less coverage than the Rotary Pruner, which had a final coverage of 0.78. Cycle 2L achieved significantly more coverage (10.7\% more) during day 50 than Cycle 2R, which could be in part due to the greater number of prunes of Cycle 2R.

In general, the Pruning Shears executed much cleaner cuts than the Rotary Pruner and sacrificed less total canopy coverage. To try to match the effectiveness of the Rotary Pruner in terms of diversity for future gardens, the Pruning Shears could make multiple cuts per plant or could prune more frequently than every 5 days.  


%% file: figures/cycle_1_diversity_and_coverage.tex
\begin{table}
\centering
\vspace{0.15cm}
\scalebox{0.85}{
\begin{tabular}{ |p{2cm}|r|r|r|r|}
 \hline
 Plant Type & $r_{max}$ & Cycle 1L & Cycle 1R & $\%$ Change \\
 \hline
 \hline
 Kale & 37 & 0.158 & 0.102 & -35.44$\%$ \\
 \hline
 Turnip & 33 & 0.085 & 0.043 & -49.41$\%$ \\
 \hline
 Borage & 32 & 0.122 & 0.076 & -37.70$\%$\\
 \hline
 Swiss Chard & 28 & 0.105 & 0.102 & -2.86$\%$\\
 \hline
 Arugula & 25 & 0.098 & 0.121 & 23.47$\%$\\
 \hline 
 Radichhio & 23 & 0.034 & 0.059 & 73.53$\%$\\
 \hline
 Red Lettuce & 20 & 0.000 & 0.057 & N/A\\
 \hline 
 Cilantro & 19 & 0.062 & 0.078 & 25.81$\%$\\
 \hline
 Green Lettuce & 16 & 0.028 & 0.095 & 239.29$\%$\\
 \hline
 Sorrel & 10 & 0.002 & 0.031 & 1450$\%$\\
 \hline
 \hline
 DIVERSITY &  & 0.856 & \textbf{0.970} & 13.32$\%$\\
 \hline
 COVERAGE & & \textbf{0.924} & 0.784 & -15.15$\%$\\
 \hline
\end{tabular}}
\caption{\textbf{Plant Type Metrics for Garden Cycles 1L \& 1R.} This table shows diversity and coverage for plant types on day 60. The values for Cycle 1L (not pruned) and Cycle 1R (pruned with Rotary Pruner) are calculated via $[c_i(60)*(R/R_i)^2]$ for each plant type (Section III). The goal of pruning is to foster a diverse garden while maintaining a high coverage.}
\squeeze
\label{table:cycle_1_div}
\end{table}

%% file: figures/cycle_2.tex
\begin{figure}
    \vspace{0.2cm}
    \centering
    \frame{\includegraphics[width=42mm, clip]{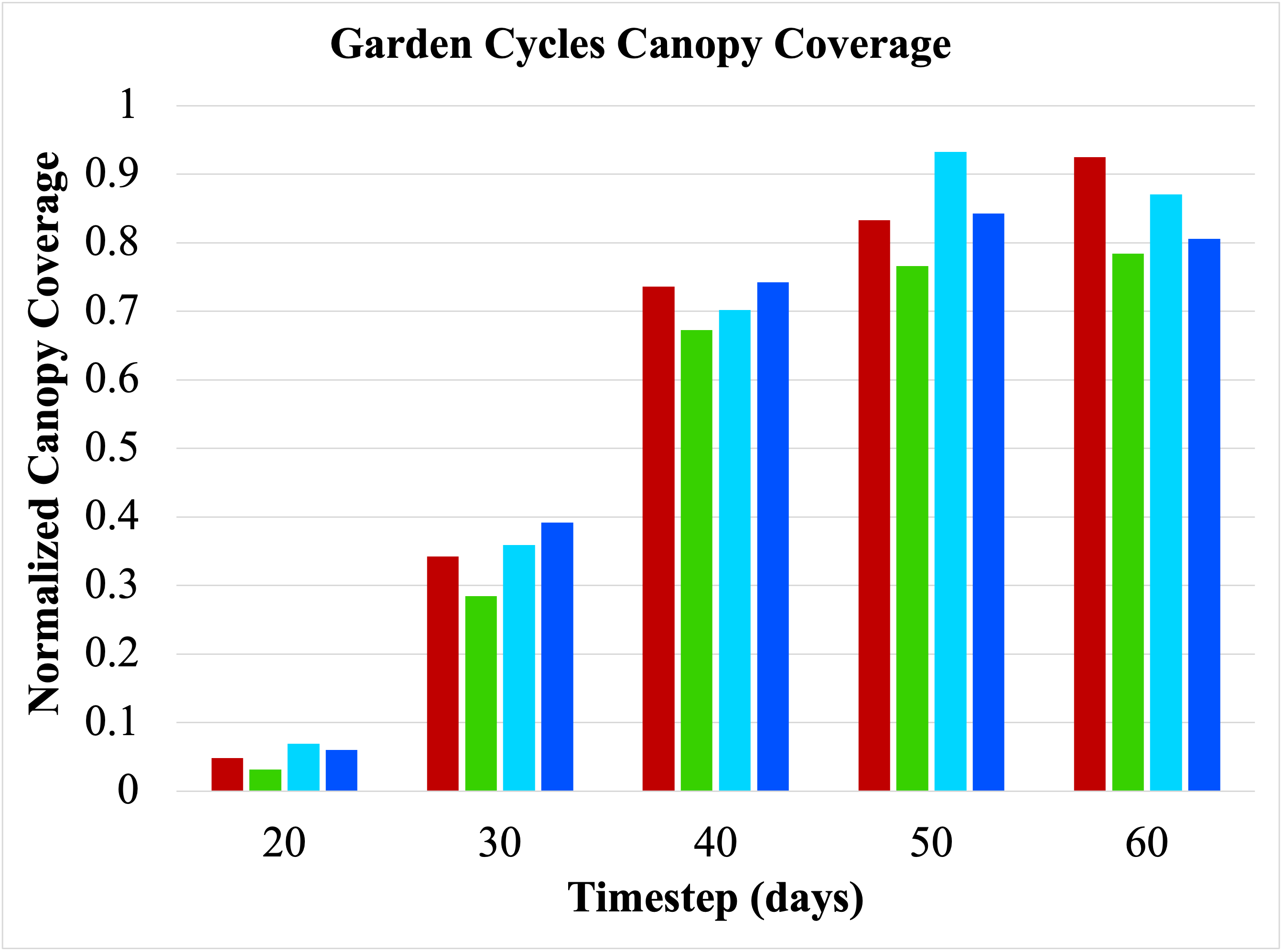}}
    \frame{\includegraphics[width=42mm]{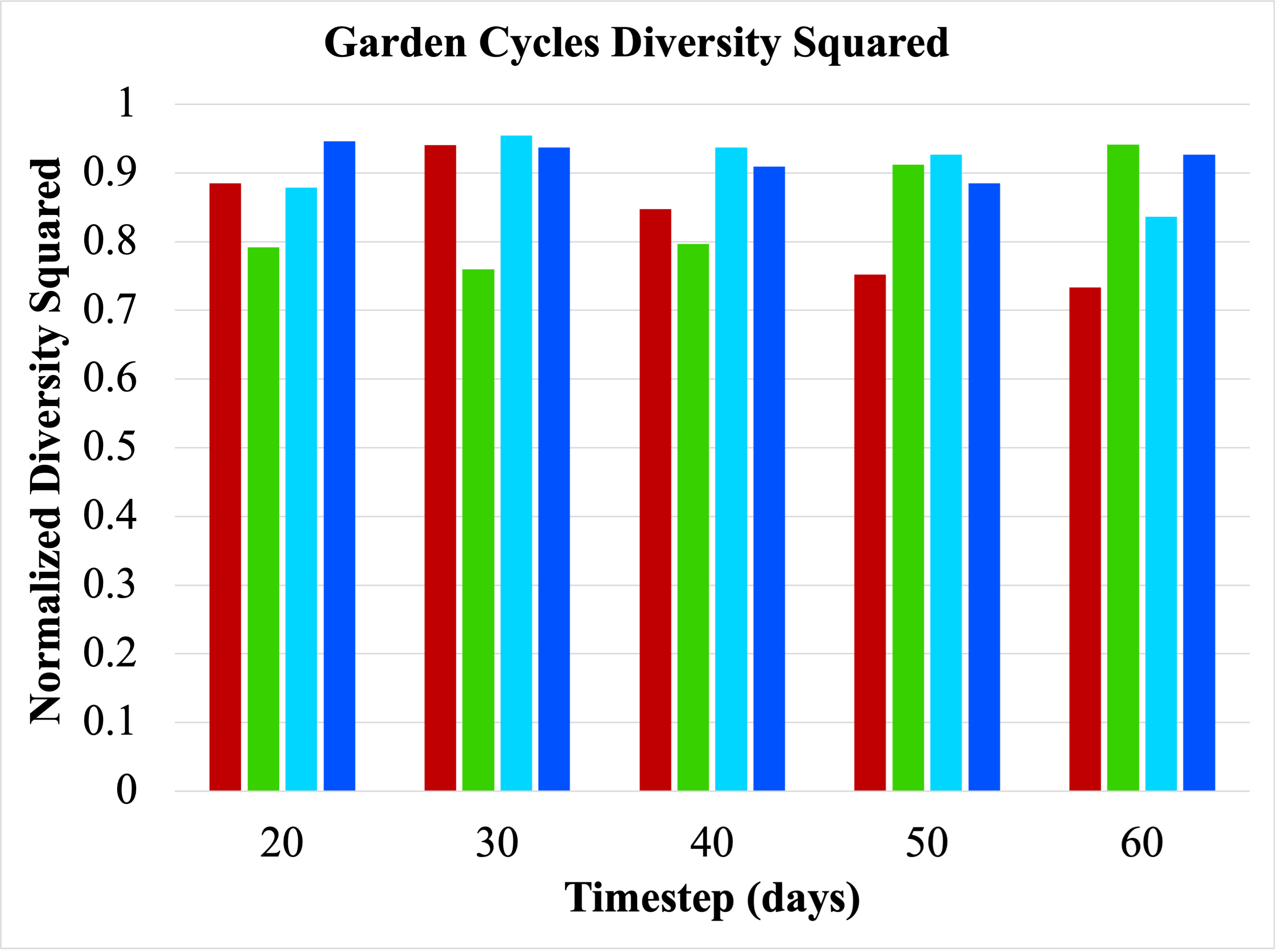}}
    \frame{\includegraphics[width=85.5mm]{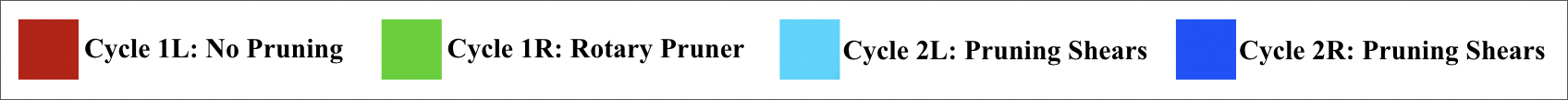}}
    \caption{\textbf{Garden Cycle Comparison.} Data points were recorded for days 20, 30, 40, 50, and 60 through hand labeled phenotyping masks. \textbf{Left:} Comparison of the coverage of the 4 Garden Cycles. The non-pruned garden has the highest value by day 60, with Cycle 2L (pruning shears) not far behind. \textbf{Right: } Comparison of the diversity squared of the 4 Garden Cycles. The non-pruned garden had lowest diversity by day 60, and Cycles 1R (rotary pruner) and 2R (pruning shears) achieved the highest diversity.} 
    \vspace{-0.6cm}
\label{fig:cycle_2}
\end{figure}

%% file: section-9-conclusion.tex
\section{Conclusion}

Despite recent advances in robotics and automation, automating a polyculture garden remains challenging. This paper presents algorithms for the automated pruning of polyculture plants to encourage plant diversity while maintaining canopy coverage. We present a learned Plant Phenotyping Network and Bounding Disk Tracking algorithm to estimate the state of a polyculture garden. For automated pruning, we present a Prune Point Identification Network and visual servoing to transfer from pixel coordinates to real world FarmBot coordinates. We designed and evaluated two custom pruning tools. We evaluated the automated pruning system in a real world polyculture setting over the course of 4 Growth Cycles. In future work, we plan to integrate policies to optimize water usage, and explore using seed placement as an inductive bias to our phenotyping model. We also plan to explore closed-loop visual servoing for the Pruning Shears.



%% file: section-10-acknowledgements.tex
\section*{Acknowledgements}
\vspace{-0.1cm}
\relsize{-1}
This research was performed at the AUTOLAB at UC Berkeley in affiliation with the Berkeley AI Research (BAIR) Lab, and the CITRIS "People and Robots" (CPAR) Initiative. This research was supported in part by the RAPID: Robot-Assisted Precision Irrigation Delivery Project (USDA 2017-67021-25925 under NSF National Robotics Initiative). We thank our colleagues who provided helpful feedback and suggestions, in particular Mary Power, Sarah Newman, Eric Siegel, Isaac Blankensmith, Maya Man, Christiane Paul, Charlie Brummer, Christina Wistrom, and Grady Pierroz.